\documentclass[letterpaper, 10 pt, conference]{ieeeconf}  
\usepackage[font=footnotesize]{caption}
\usepackage[font=footnotesize]{subcaption}

\IEEEoverridecommandlockouts                              

\usepackage{amsmath} 
\usepackage{amssymb}  
\usepackage{bm}
\usepackage[linesnumbered,ruled]{algorithm2e}
\usepackage[colorinlistoftodos,disable]{todonotes}
\usepackage{hyperref}

\usepackage{microtype}
\usepackage{caption}
\usepackage{subcaption}
\usepackage{graphicx}
\usepackage{placeins}
\usepackage{multirow}
\title{\LARGE \bf
Using Local Experiences for Global Motion Planning
}

\author{Constantinos Chamzas$^{1}$, Anshumali Shrivastava$^{1}$,  Lydia E. Kavraki$^{1}$ 
\thanks{*This work has been supported in part by NSF 1718478 and Rice University Funds.}
\thanks{$^{1}$ Department of Computer Science, Rice University, Houston TX, USA 
{\tt\small \{chamzas, anshumali, kavraki\}@rice.edu}}%
}
\begin{document}

\maketitle
\thispagestyle{empty}
\pagestyle{empty}

\begin{abstract}
    Sampling-based planners are effective in many real-world applications such as robotics manipulation, navigation, and even protein modeling.
    However, it is often challenging to generate a collision-free path in environments where key areas are hard to sample.
    In the absence of any prior information, sampling-based planners are forced to explore uniformly or heuristically, which can lead to degraded performance.
    One way to improve performance is to use prior knowledge of environments to adapt the sampling strategy to the problem at hand.
    In this work, we decompose the workspace into local primitives, memorizing local experiences by these primitives in the form of local samplers, and store them in a database.
    We synthesize an efficient global sampler by retrieving local experiences relevant to the given situation.
    Our method transfers knowledge effectively between diverse environments that share local primitives and speeds up the performance dramatically.
    Our results show, in terms of solution time, an improvement of multiple orders of magnitude in two traditionally challenging high-dimensional problems compared to state-of-the-art approaches.
\end{abstract}

\section{INTRODUCTION}

Motion planning is an integral part of many areas of robotics.
Robots operating autonomously need to generate many
different motion plans in complex environments.
This is true especially in the context of task and motion planning~\cite{Dantam2018}. Even a single task, such as stacking blocks, might require querying a motion planner thousands of times.
Humans can execute motions instantaneously that robots currently struggle with. To achieve human-level behavior,  fast online motion planning is essential. 

The widespread success of sampling-based planners lies in their ability to approximate the connectivity of high-dimensional spaces with a small number of samples~\cite{Choset2005}.
However, in many cases regions necessary for connectivity are unlikely to be sampled by an uninformed sampler.
This is known as the  \textit{narrow passages} problem~\cite{Hsu2003a,Hsu1998,Amato1998} and greatly limits the performance of sampling-based planners in many scenarios.

\begin{figure} \centering
    \includegraphics[width =1\linewidth]{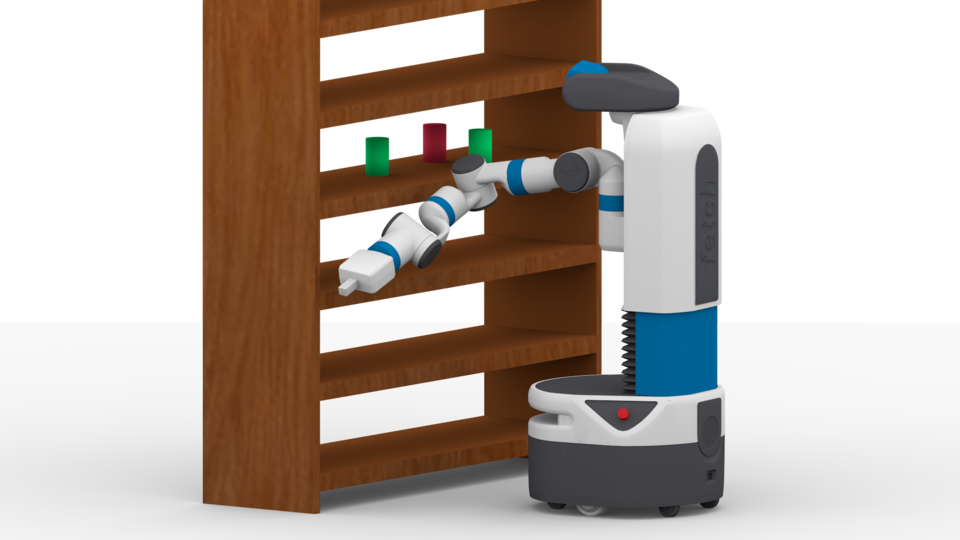}
    \caption{The task of grasping the red can without removing the green cans is very challenging for traditional methods. The proposed approach efficiently solves it by using prior experiences to guide the search.}
    \label{fig:fetch_red}
\end{figure}

Among the possible approaches to solving problems that involve narrow passages is the emerging field of \mbox{experience-based} planning~\cite{Coleman2015,Ichter2018,Phillips2012}.
It is common for robots during their operation to come across similar workspaces resulting in similar motion plans.
Such a case can be seen in~\autoref{fig:fetch_red} where a robot needs to grasp the red can and put it on the top shelf. 
In this case, prior knowledge about similar scenarios could expedite the motion planning process. 
By biasing sampling towards interesting regions~\cite{Ichter2018} or by retrieving and reusing old solutions~\cite{Coleman2015}, experience-based methods try to transfer knowledge obtained from similar problems to others.
Unfortunately, small changes in the workspace can drastically affect the possible solutions, in several cases, thus making generalization difficult~\cite{Kim2017c,Lehner2018}.

This paper presents a sampling strategy for sampling-based planners that aids in discovering the connectivity of the configuration space even for pathological cases.
This sampling strategy utilizes a decomposition of the workspace into local primitives.
The main insight of our method is that learning to generate important samples in the configuration spaces defined by the robot and the primitives helps approximate the configuration space of the global workspace.
This approach can generalize to new environments that contain the workspace primitives used earlier or primitives very similar to them.
In this work, we focus on problems with simple geometric features, yet manage to solve a class of problems that were practically unsolvable by modern sampling-based motion planners.
Also, we raise the question of whether the notion of decomposition applies in unstructured environments with complex geometries, tackling problems that were previously beyond the reach of sampling-based motion planners.

The contributions of this paper are threefold.
First, we propose the decomposition of the workspace into local primitives, and we solve motion planning problems in workspaces that contain only the local primitives.
The result of this step is the estimation of local samplers that produce samples in the difficult regions of the configuration spaces of the local primitives. The parameters of these  local samplers are stored in a database. Second, we show how to synthesize a global sampling strategy based on these local samplers.
Third, we show the effectiveness of our approach in two challenging environments where it achieves significant improvement over existing methods.
\section{BACKGROUND}

Sampling-based planners have been widely adopted in robotics due to their ability to scale to high-dimensional problems.
The two main categories are graph-based approaches (e.g, \textsc{prm}~\cite{Kavraki1996}) for \mbox{multi-query} problems and \mbox{tree-based} (e.g, \textsc{rrt}~\cite{Kuffner2000},~\textsc{est}~\cite{Hsu})  for \mbox{single-query} problems.
Although in general motion planning is \textsc{pspace}-complete~\cite{Canny1988,Canny1988b}, sampling-algorithms perform remarkably well.
However, in challenging environments (e.g., instances with narrow passages), sampling plays an increasingly important role in the planning performance.
It is theoretically understood, however, that using \mbox{non-uniform} samplers to generate more samples in \mbox{low-expansiveness} areas~\cite{Hsu} can alleviate this problem. 

Many approaches use rejection sampling, where the sample is accepted only if it passes a specific geometric test such as
\mbox{Bridge-Sampling}~\cite{Hsu2003a} or \mbox{Gaussian-Sampling}~\cite{Boor1999a}. Other approaches try to guess difficult areas, such as \mbox{Medial-Axis} sampling~\cite{Brock2004} or \mbox{Obstacle-Based} Sampling~\cite{Amato1998}. 
Although these approaches ultimately generate samples near or inside narrow passages, they still consider the entire configuration space which is computationally expensive.
Nevertheless, the resulting graph is typically much smaller than the one produced by uniform sampling.

Recent approaches similar to our method try to leverage acquired information about the problem, to bias the sampling.
Reinforcement learning was used by~\cite{Zucker2008}  to infer important areas of the workspace that were transformed to configuration samples.
The authors of~\cite{Ichter2018} utilize a generative model, called Conditional Variational \mbox{Auto-Encoder} (\textsc{cvae}), that learns to produce samples that lie in ``interesting'' areas given a workspace description.
\mbox{Sampling-biasing} methods have the advantage that they can be used with many sampling-based planners without any modification.
However, both of these methods rely on a model that uses workspace features to infer important samples in the configuration space.
In~\cite{Zucker2008}, a discrete workspace cell was mapped to a configuration through the inverse kinematics of the \mbox{end-effector}. 
In~\cite{Ichter2018}, a neural network was used to infer these samples.
Our experiments showed that in complex configuration spaces these models do not consistently produce samples in the important areas of the configuration space. 

\mbox{Online-adaptive} sampling methods use collision checking information to infer which areas of the configuration space are important at runtime.
\mbox{Utility-guided} sampling~\cite{Burns2005} chooses samples with the maximum information gain based on the entropy of the roadmap.
The authors of~\cite{Arslan2015} formulate the sampling problem as a classification between free and \mbox{in-collision} samples. 
\mbox{Toggle-\textsc{prm}} \cite{Denny2013} creates two roadmaps, one in the configuration space and one in the obstacle space, and tries to infer samples in the narrow passages.
These methods adapt the sampling online, based on the state the motion planning algorithm, but do not transfer this knowledge between different planning queries.
On the contrary, the proposed method biases the sampler based on previous planning queries. Thus, \mbox{online-adaptive} sampling can be used in parallel with the proposed sampling biasing.

Orthogonal to these methods are database approaches. 
Instead of modifying sampling, these methods leverage previous experiences by storing discrete paths or graphs in a database.
This information is later retrieved and repaired/transformed to satisfy the new kinematic constraints.
These methods can be thought of as \mbox{hard-coded} experiences compared to biasing the sampling.

The authors of~\cite{Berenson2012a}  used a library of paths that was queried based on the proximity of \mbox{start-goal} configurations of the entry and the query.
If a valid path could not be retrieved based on these heuristics, \textsc{birrt}~\cite{ Kuffner2000} was used to repair the invalid parts.
This idea was expanded in~\cite{Coleman2015} where a graph was used to store the paths removing redundancy.
Although improving the execution time by orders of magnitude, the mentioned approaches do not adapt well to changes and yield improved results mainly in invariant environments.
This happens because they do not explicitly include the workspace in their experience representation.

Database approaches that explicitly use workspace information include~\cite{Lien2009} which creates a small database of obstacle \mbox{road-maps} smartly decomposing the configuration space in \mbox{obstacle-maps}. The trajectory prediction proposed by~\cite{Jetchev2013} saves the generated paths in \mbox{task-space} and during execution transforms them back to the configuration space by optimizing the cost of the trajectory while using IK.
Although these methods can deal with different environments~\cite{Lien2009} works only for \mbox{free-flying} robots and ~\cite{Jetchev2013} works with trajectory optimization planners that lack the probabilistic guarantees of sampling-based planners.

In this work, we combine the best of both worlds by integrating a \mbox{biased-sampler} with a database.
This is achieved by decomposing the workspace in local primitives and storing in a database the parameters of an efficient local sampler for each local primitive.  
The \mbox{biased-sampler} uses prior knowledge in a ``soft way'' avoiding the \mbox{hard-commitment} to complete paths, induced by databases methods, which may need costly repairing when there are significant changes in the workspace.
On the other hand, the \mbox{database-scheme} enables the instant mapping of local samplers to local primitives avoiding the need for a complex parametric model.        
Additionally, a database has the inherent capability of incrementally improving its experiences by simply adding new entries. The aforementioned sampling-biasing approaches would need to be retrained.

\section{METHOD OVERVIEW}
\begin{figure*}
    \centering
    \includegraphics[width = \linewidth]{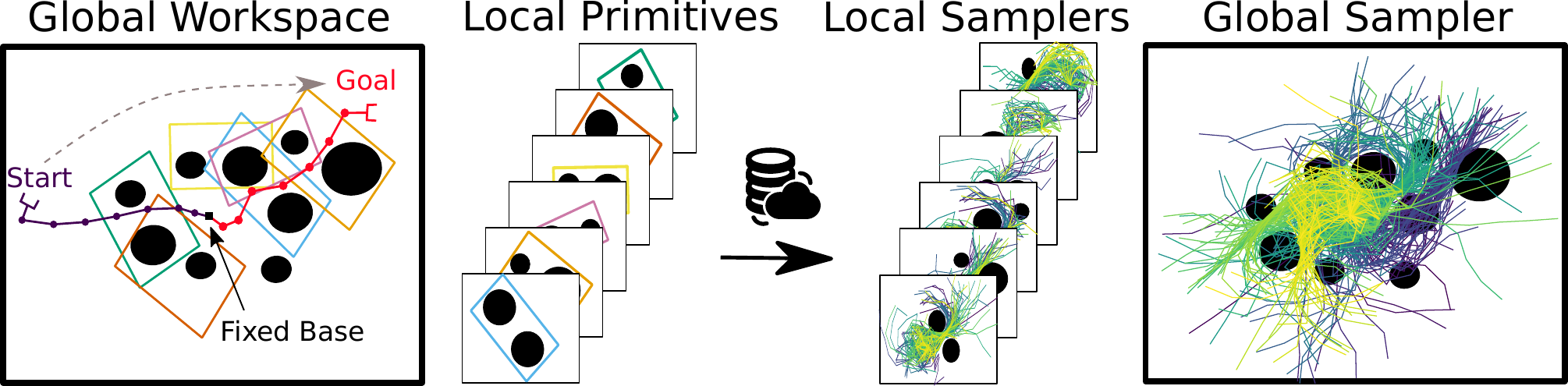}
    \caption{An illustration of our proposed method. In this example, the workspace contains circular obstacles and a planar arm with a fixed base. In our approach, the workspace is decomposed to local primitives. During the offline phase, the local samplers are computed and saved in the database. During the online phase, each local primitive is mapped to its corresponding local sampler through the database. Finally, the local samplers are synthesized to a global sampler.}
    \label{fig:pipeline}
\end{figure*}
\vspace{1pt}

In this work, we modify the sampling step that produces configuration samples,  which is at the core of all sampling-based planners as shown in~\autoref{alg:1}. Similar to~\cite{Ichter2018} we generate $\lambda N$ samples from the global sampler (\autoref{alg1:G}) and $(1-\lambda)N$ from  the uniform sampler (\autoref{alg1:U}). $\lambda$ is a hyperparameter that is determined by the application. The data structure G and the update() function (\autoref{alg1:l3}) differentiates the sampling-based planners.
For example in a \mbox{\textsc{prm}-like} setting, G is a roadmap, and new samples are added to expand the roadmap which captures the connectivity of the \mbox{C-Space} in a variety of ways ~\cite{Kavraki1996,Denny2013}.  
In an \mbox{\textsc{rrt}-like} setting, G is a tree, and the sample is used to expand the tree in different ways~\cite{Kuffner2000,Denny2013}.
However, as stated above, regardless of the planner details, the performance can be improved by using informed sampling.    
\begin{algorithm}
    \SetKwInOut{Input}{Input}
    \SetKwInOut{Output}{Output}
    
    \Input{Number of iterations $N$}
    \Output{Graph structure $G$ }
    \While {i $\leq$ N ~$or$~ solutionFound()}
    {
        \eIf {rand(0,1) $\le \lambda$}
        {
            x $\sim$ \textbf{GL-Sampler()} \label{alg1:G}
        }
        {
            x $\sim$ Uniform()  \label{alg1:U}
        }
        update(G, x)\label{alg1:l3}
    }
    \Return G
    \caption{Modified Sampling Procedure}
    \label{alg:1}        
\end{algorithm}

The proposed sampling strategy is based on local primitives of the workspace.
The key insight of our approach is that different regions of the workspace often create relatively uncorrelated narrow passages in the configuration space.
For example, the top and bottom shelf of a bookcase will create two different challenging regions that the robot arm will need to traverse when moving from one shelf to the other. This means that we can bind an effective local sampling distribution to each shelf (local primitive) dealing with the two problems independently. By combining the local samplers, we can synthesize a global sampler that informatively guides the planners.  Our experiments showed that even in highly correlated cases, our approach is still effective.

We will introduce some notation, before describing the main algorithm.
Since our method relies on a workspace description, we will denote the workspace as ${W=\{wo_1,\ldots,wo_M \}}$ where each $wo_i$ is a workspace obstacle.
Also we denote the set of local primitives as ${LW = \{lw_1, \ldots , lw_N\}}$, such that  $ \bigcup_{i=1}^{i=N} lw_i\subseteq W$.
Note that the local primitives do not have to be mutually exclusive.
For example, pairs of workspace obstacles ${lw_i = \{wo_j, wo_k\}}$ could be a valid set of local primitives.
The global target distribution is denoted as $p(x|W)$.
Local samplers that are used to approximate it are denoted as  $p_i(x|lw_i)$.

The main steps of the Global-Local Sampler (GL-Sampler) are outlined in~\autoref{alg:2}. 
The first step~(\autoref{alg2:decomp}) is to decompose the workspace $W$ by identifying its local primitives $LW$.
In general, this could be a highly sophisticated function; however, in the context of this work, the local primitives are always pairs of workspace obstacles ${lw_i=\{wo_j,wo_k\}}$.
For each of the local primitives, we try to retrieve their corresponding samplers~(\autoref{alg2:ret2}) based on a similarity function $k$~(\autoref{alg2:ret1}). 
If no similar local primitive exist in the database, then the parameters of the local sampler are calculated~(\autoref{alg2:create1}) and stored in the database~(\autoref{alg2:create2}).
Note that the database can also be computed offline if the possible local primitives are known \mbox{before-hand} which is often the case.
The local samplers are combined to synthesize the global sampler~(\autoref{alg2:create3}). The GL-Sampler is created only at the beginning of the motion planning query.

\begin{algorithm}
    \SetKwInOut{Input}{Input}
    \SetKwInOut{Output}{Output}
    
    \Input{workspace $W$, database $D$, threshold d}
    \Output{GL-Sampler()}
    Decompose $W$ to  $LW = \{lw_{1}\ldots lw_{N}\}$ \label{alg2:decomp} \\
    \ForEach{ $lw_i \in LW$ }{
        \eIf {$\exists lw_j \in D $~s.t.~ $k(lw_j,lw_i) <d$ \label{alg2:ret1}}
        { 
            Retrieve $p_i(x|lw_i)$ from $D(lw_i)$\label{alg2:ret2}
        }
        {
            Create $p_i(x|lw_i)$ \\ \label{alg2:create1} 
            Insert $lw_i:p_i(x|lw_i)$ in $D$\label{alg2:create2}
        }
    }
    Synthesize $\hat{p}(x|W)$ from $p_i(x|lw_i)$ \\\label{alg2:create3}  
    
    \Return {$\hat{p}(x|W)$} 
    \caption{Create GL-Sampler}
    \label{alg:2}        
\end{algorithm}

\autoref{fig:pipeline} shows an illustrating example of the algorithm where it is applied on a fixed-base, \mbox{8-link} planar manipulator (\mbox{non-intersecting}) in an environment of circular obstacles. 
The local primitives in this case are pairs of circles and are described as $lw_{i} =\{ x_a, y_a, r_a, x_b, y_b, r_b\}$,  where $x,y,r$ denote the position and radius of each circle.
In the following sections, the creation of the database, retrieval of local sampler and composition of the global sampler are described. 

\subsection{Creating the Database of Local Samplers}
\label{sec:creation}

Each local sampler $p_i(x|lw_i)$ must produce samples that quickly capture the local primitive's configuration space.
For example, each local sampler should produce samples in narrow passages of the corresponding \mbox{C-Space}.
First, we generate such samples and later fit the local sampler to them.
Such samples are created by solving motion planning queries that likely traverse difficult regions of the configuration space of the local primitives.
For every local primitive we \mbox{pre-specify} a set of such motion planning queries.
For the local primitives (pairs of circles) in~\autoref{fig:pipeline}, a path that starts with the robot between the circles and ends with the robot entirely out of them likely traverses a narrow passage.  
A standard sampling planner e.g., \textsc{rrt}, \textsc{birrt}~\cite{Kuffner2000} is used to solve these queries quickly. 
Additionally, it is imperative that multiple paths for the same query are generated to be robust to obstacles that are part of the global workspace but not the local primitive.
This is clear in \autoref{fig:pipeline} where several samples which were valid for the local primitives are invalid for the global problem. 
To deal with this we run the chosen planner multiple times, creating different paths due to its randomness.
Also by using shortcutting techniques ~\cite{Geraerts2007}, most of the redundant samples can be removed to increase the ratio of the useful samples. 

Due to the complexity of the needed distribution and its natural \mbox{multi-modality}, we choose a Gaussian Mixture Model (\textsc{gmm}) as the local sampler similar to \cite{Lehner2018}.
However, contrary to~\cite{Lehner2018} we do not use the traditional expectation-maximization algorithm to calculate the parameters of the \textsc{gmm}.
There is no good way to choose the number of mixtures, and more importantly, the distance between configurations does not necessarily relate to \mbox{C-Space} connectivity which is what sampling-based algorithms need to capture.
Instead, for the local sampler $i$ we choose to place one mixture to each produced configuration ${q_{i1}, \ldots, q_{i\textsc{M}}}$ and use a fixed covariance $\Sigma_i = \sigma I$ where $\sigma$ is a hyperparameter and $I$ is the identity matrix.
This might create an unnecessarily large amount of mixtures, but we present a way to reduce them while respecting the connectivity in \autoref{sec:transform}.
The local sampler will be:

\begin{align}
\label{eq:1}
{p}_i(x|lw_i) =\textsc{gmm}(\mu,\Sigma)  = \frac{1}{M_i}\sum_{j=1}^{M_i} \mathcal{N}(q_{ij},\Sigma_{ij})
\end{align}

$M_i$ is the number of mixtures. We choose a uniform vector for the weights making all the mixtures equiprobable. In the database, we save the parameters of this distribution and its corresponding local primitive.

\subsection{Retrieving Local Samplers}
To retrieve the local samplers from the database, we need a similarity function $k(lw_i,lw_k)$ to compare the local primitives between them.
General workspace descriptors and possible similarity functions have been described by \cite{Jetchev2013}.
In our case where the primitives are simple geometric descriptions the negative squared Euclidean distance is used: 
\begin{align*}
k(lw_i,lw_j) =  -(lw_i - lw_j)^T(lp_i - lw_j)
\end{align*}

We retrieve all the parameters of local samplers that are above a certain threshold $d$ of this similarity.
Since the local samplers retrieved will not correspond to the exact local primitives a \textit{similarity error} is introduced.  

\subsection{Synthesizing the Global Sampler}

Now we will describe how the local samplers approximate the global sampler.
Given an arbitrary partition of the configuration space $\bigcup^N_{i=1} X_i = X$,~$ X_i \bigcap X_j= \emptyset$, ${ \forall i,j,~ i\neq j} $, the global sampler can be expressed as a sum of other distributions using the law of total probability:
\begin{align}\label{eq:2}
p(x|W) = \sum_{i=1}^N p(x| W, X_i) p(X_i)  = \sum_{i=1}^N p_i^{\star}(x| W)a_i 
\end{align}

In the last equation we rewrote  $p(X_i)$ as $a_i$ and $p(x| W, X_i)$ as $p^{\star}_i(x|W)$. Note that the support of $p^{\star}_i(x|W)$ is $X_i$.
We approximate it in the following way:
\begin{align}\label{eq:3}
p^{\star}_i(x| W)  \approx  p_i(x| W)\approx  p_i(x| lw_1, \ldots ,lw_N) \approx p_i(x| lw_i) 
\end{align}

Three approximations are used in the derivation above.
The first is one is that $p_i(x|W)$ has its support on $X$ instead of $X_i$.
This induces only a small error because $ p_i(x| lw_i)$  is a distribution that has values close to zero outside $X_i$. The second one is that most of the information in  $W$ is incorporated in the set of local primitives $\{lw_1 , \ldots lw_N \}$ which is true if the local primitives are responsible for most of the difficult regions of the configuration space. The final approximation is that the local primitives independently affect only one local distribution. ${p_i(x| lw_1, lw_2, \ldots ,lw_N) \approx p_i(x|lw_i)}$.
This is not true in general especially in cases where the local primitives are close together. This is the reason why multiple paths are created for each local primitive in \autoref{sec:creation}. We refer to this as the \textit{decomposition error}. In the experiments section, we show empirically that even when this error is large our sampling method is much more effective than uniform random sampling. Finally combining \autoref{eq:1}, \autoref{eq:2} and \autoref{eq:3} the global sampler is:

\begin{align*}
p(x| W) \approx  \sum_{i=1}^N \frac{a_i }{M_i}\sum_{j=1}^{M_i}  \mathcal{N}(q_{ij},\Sigma_{ij}) =
\sum_{i=1}^N \frac{1}{N} \sum_{j=1}^{M_i} \mathcal{N}(q_{ij},\Sigma_{ij})
\end{align*}

We set  $a_i = \frac{M_i}{N}$ which means that the weight of each local sampler is proportional to the number of its mixtures.

\section{Reducing the Database}
\label{sec:reduce}
Since querying the database happens online, it is crucial to keep its size at a minimum for fast retrieval.
We propose two such reductions.
One is removing mixtures from the database by merging cliques, and the other is transforming the local experiences to account for multiple local primitives.
\subsection{Merging Cliques}
\begin{figure}
    \begin{subfigure}[b]{0.32\columnwidth}
        \includegraphics[width=\linewidth]{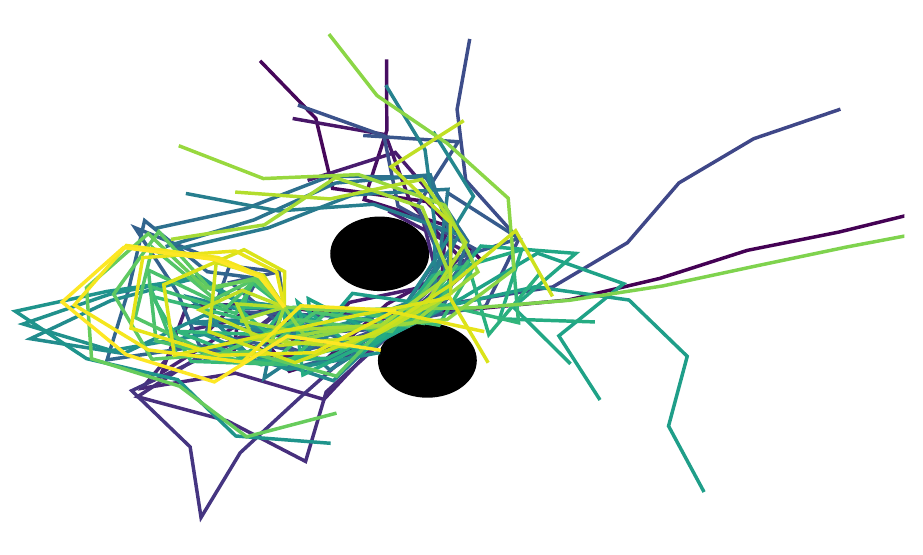}
        \caption{~}
        \label{fig:r1}
    \end{subfigure}
    \begin{subfigure}[b]{0.32\columnwidth}
        \includegraphics[width=\linewidth]{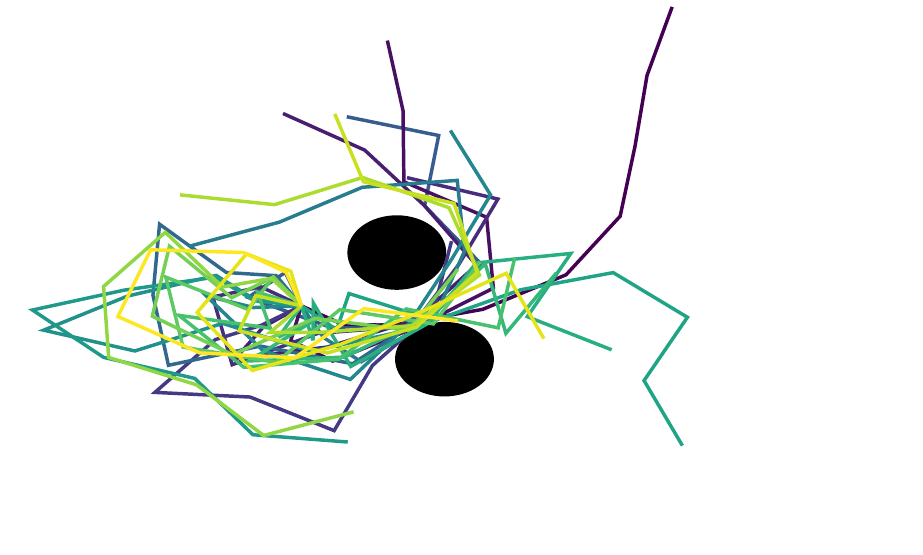}
        \caption{\hspace*{1em}} 
        \label{fig:r2}
    \end{subfigure}
    \begin{subfigure}[b]{0.32\columnwidth}
        \includegraphics[width=0.6\linewidth]{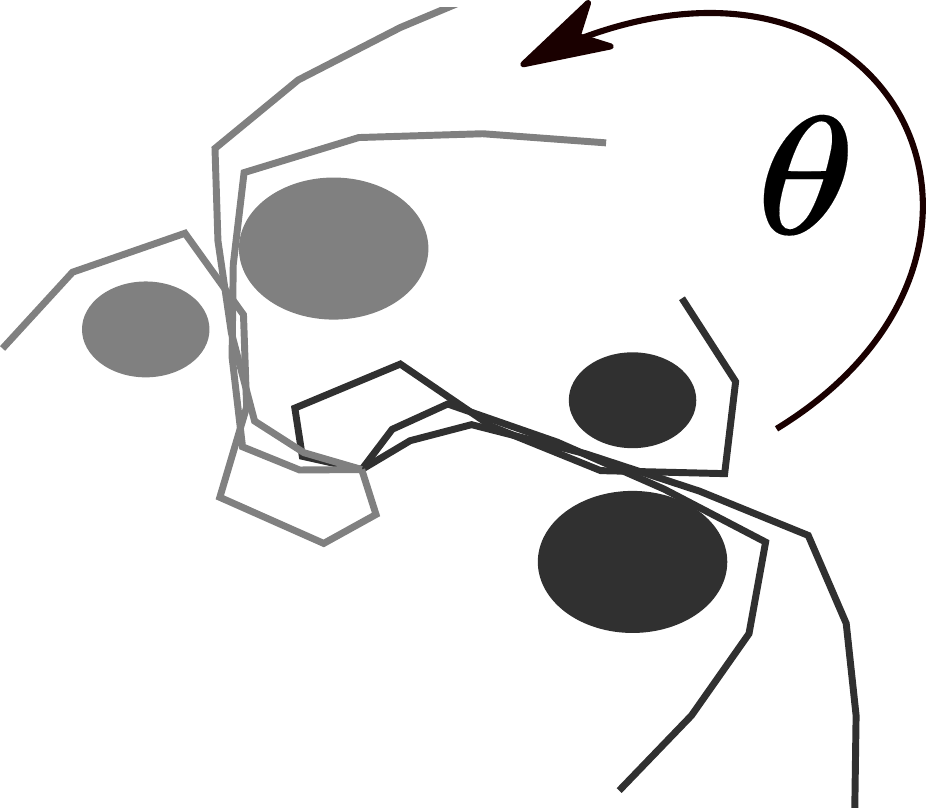}
        \caption{\hspace*{2em}} 
        \label{fig:r3}
    \end{subfigure}
    \caption{(a) \mbox{Un-merged} Cliques, 53 mixtures  needed (b)~Merged Cliques,  23 mixtures  needed, (c)~Transforming the local sampler of the black primitive to apply to they grey primitive.}
\end{figure}

This step can be executed after the generation of the useful configurations in \autoref{sec:creation}.
Using the extracted configurations ${q_{i1}, \ldots, q_{i\textsc{m}}}$ for a specific local primitive $lw_i$ we create a graph by connecting two configurations with an edge if there is a \mbox{collision-free} straight line between them.
From this graph, we identify the fully connected subgraphs, also known as cliques.
The cliques essentially represent groups of configurations that can be accurately approximated with one mixture model.
We are not interested in finding the optimal number of cliques which is an \textsc{np-hard} problem, and for that reason, we are using a greedy algorithm that finds them sequentially.
After finding the cliques, we calculate the mean and the covariance (if enough samples exist) of each one and use them as a parameter to the \textsc{gmm} as in \autoref{eq:1}.
Our experiments showed that the time performance was similar when using the merged or the \mbox{un-merged} cliques. 
An example before merging the cliques can be seen in \autoref{fig:r1} where the \textsc{gmm} model would have 53 mixtures, after merging \autoref{fig:r2} shows the \textsc{gmm} would have 23 mixtures which save a lot of space in the database.

\subsection{Transforming the Local Experiences}

\label{sec:transform}
To reduce the size of the database,  we employ a transformation scheme similar to \cite{Lien2009} that takes advantage of the inherent symmetries of the robot.
This allows a local sampler $p_i(x|lw_i)$ to be transformed to $p_i'(x|lw_i')$ where $lw_i'$ is a transformed local primitive.
Essentially the inherent symmetries of a robot are changes in the configuration of the robot that counter transformations of the local primitives like rotation or translation.
Although this is not true in the general case, the majority of robots such as mobile manipulators, drones, and robotic arms have such symmetries.   
In the kinematic chain scenario \autoref{fig:pipeline}, the inherent symmetry is its rotational invariance around its fixed base.
This is illustrated in \autoref{fig:r3} where the grey local primitive is the rotated version of the black local primitive around the fixed base $lw_{grey} = lw_{black}*Rot(\theta)$.
Applying this simple transformation to the first joint in the means of the \textsc{gmm} $ \mu^{j1}_{grey} = \mu^{j1}_{black} +\theta$ the local distribution applies  to the new local primitive.   

This simple transformation significantly reduces storage requirements. In the studied example the dimensionality of $lw_{i} =\{ x_a, y_a, r_a, x_b, y_b, r_b\}$ is 6D and the database must store representative local primitives/local sampler from this 6D space. However, by using the mentioned transformation, this space effectively becomes 5D, since we have rotational invariance of the local primitives around the fixed base. 

\section{EXPERIMENTS}
In the following experiments, we compared different methods for the sampling part of three of the most representative sampling based planners,  \textsc{rrt}, \textsc{birrt}~\cite{Kuffner2000}, and \textsc{prm}~\cite{Kavraki1996}. We benchmarked against uniform sampling, and the Conditional Variational \mbox{Auto-Encoder} (\textsc{cvae}) proposed by~\cite{Ichter2018}. The \textsc{cvae} method is a neural network that is trained to produce samples that lie on the optimal path given a workspace description. To make the benchmarking fair we trained the \textsc{cvae} using the same dataset that was used for the estimation of the \textsc{gmm}s. We used the \textsc{ompl}~\cite{sucan2012}~benchmarking tools~\cite{moll2015} in their default settings, and run on an Intel i7 Linux machine with 4 4GHz cores and 16GB of RAM. Each query was repeated  20 times, and the timeout was set at 200 seconds. In the figures where \textsc{rrt} is not shown it did timeout in all queries. Also, the uniform to biased ratio was chosen to be $\lambda=0.5$, and the variance parameter was chosen to be $\sigma = 0.1$. 

\subsection{8-link Kinematic Chain}
Similar to~\cite{Zucker2008} we used a \mbox{fixed-based} planer arm in an environment where obstacles are of varying sizes to demonstrate the strength of our approach. The kinematic chain had 8 links of variable lengths from 1 to 2 units and the circles variable radiuses from 1 to 2 units as well. The gap between most of the circles was less than 1 unit making it a very difficult problem. As local primitives, we consider only pairs of circles that are close together. Examples of such local primitives can be seen inside the colored rectangles of \autoref{fig:pipeline}. We \mbox{pre-computed} the database such that any local primitive could be queried with a similarity error less than 3. Utilizing the transformation described in \autoref{sec:transform} only 800 pairs of obstacles were needed resulting in a small database and a very fast retrieval time (a few milliseconds). The total time for computing the database was around 10 minutes. We tested our method in three scenarios of scaling difficulty. The start configuration is noted with blue, and the goal configuration is noted with red. Note that we use log-scale for the time axis in all figures except~\autoref{fig:sce3}.
\subsubsection{Scenario1}
\begin{figure}
    \begin{subfigure}[b]{0.49\columnwidth}
        \includegraphics[width=\linewidth]{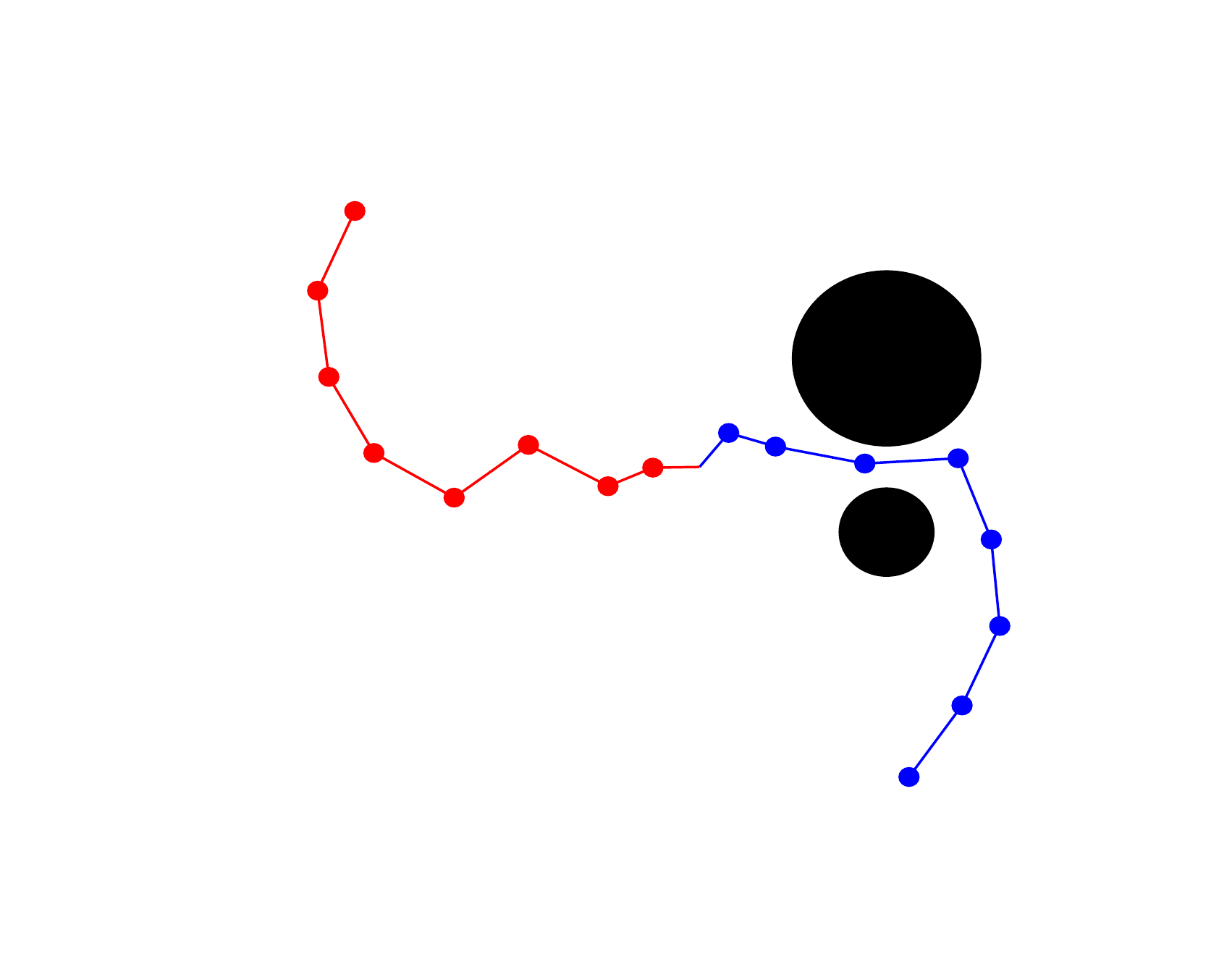}
        
    \end{subfigure}
    \begin{subfigure}[b]{0.49\columnwidth}
        \includegraphics[width=\linewidth]{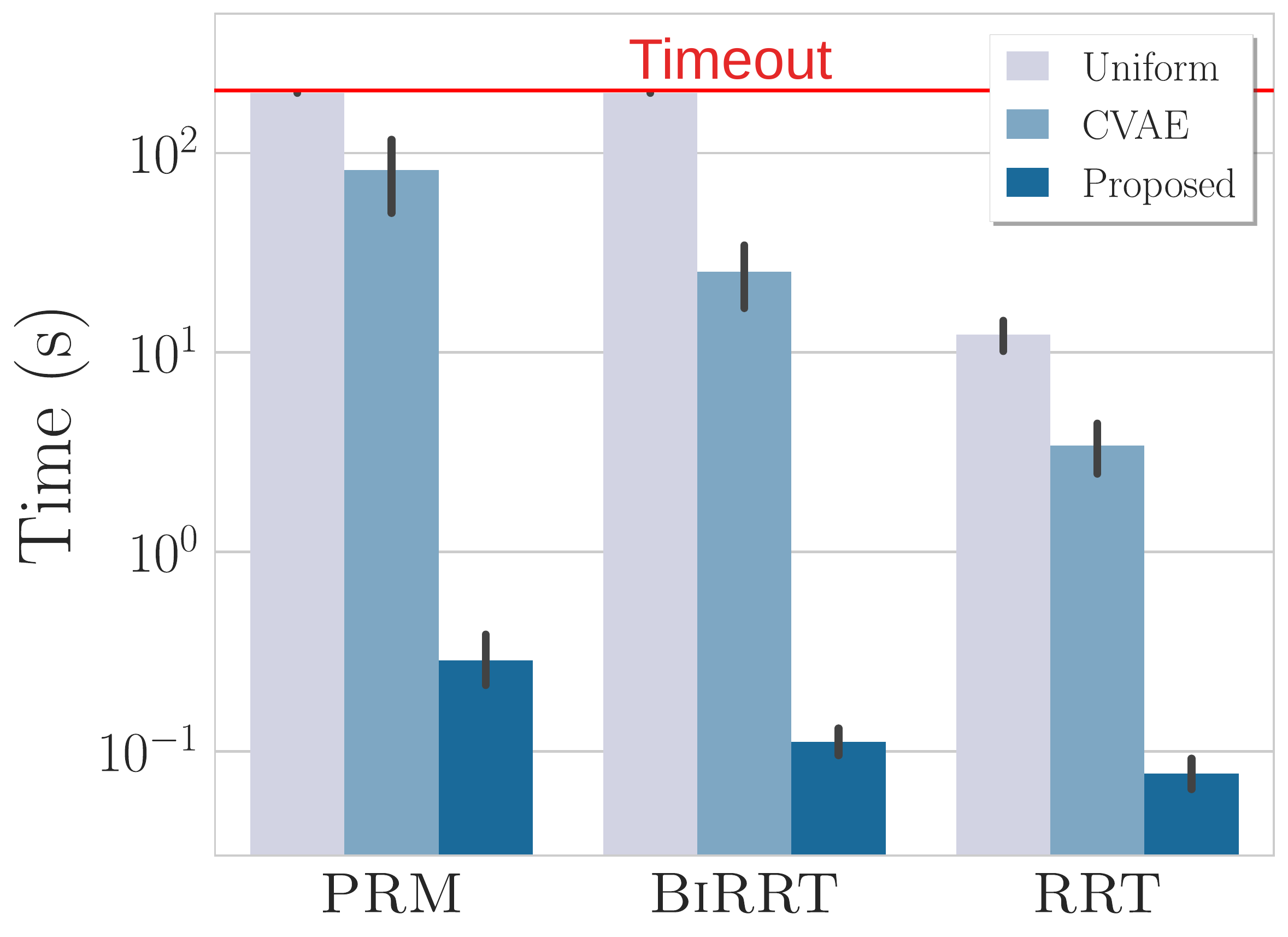}
        
    \end{subfigure}
    \caption{Scenario with low decomposition and similarity error.}
    \label{fig:sce1}
\end{figure}
The first scenario \autoref{fig:sce1}, contains only one local primitive thus having a \textit{decomposition error} of zero and a low \textit{similarity error} making it ideal for our method which found all the solutions in a fraction of a second. However, the trained \textsc{cvae} did not succeed in approximating an efficient local sampler requiring an order of magnitude more time for the same problems. Finally, it can be seen that both \textsc{prm} and \textsc{birrt} with uniform sampling did not solve any problems within 200s.
\subsubsection{Scenario2}
\begin{figure}
    
    \begin{subfigure}[b]{0.49\columnwidth}
        \includegraphics[width=\linewidth]{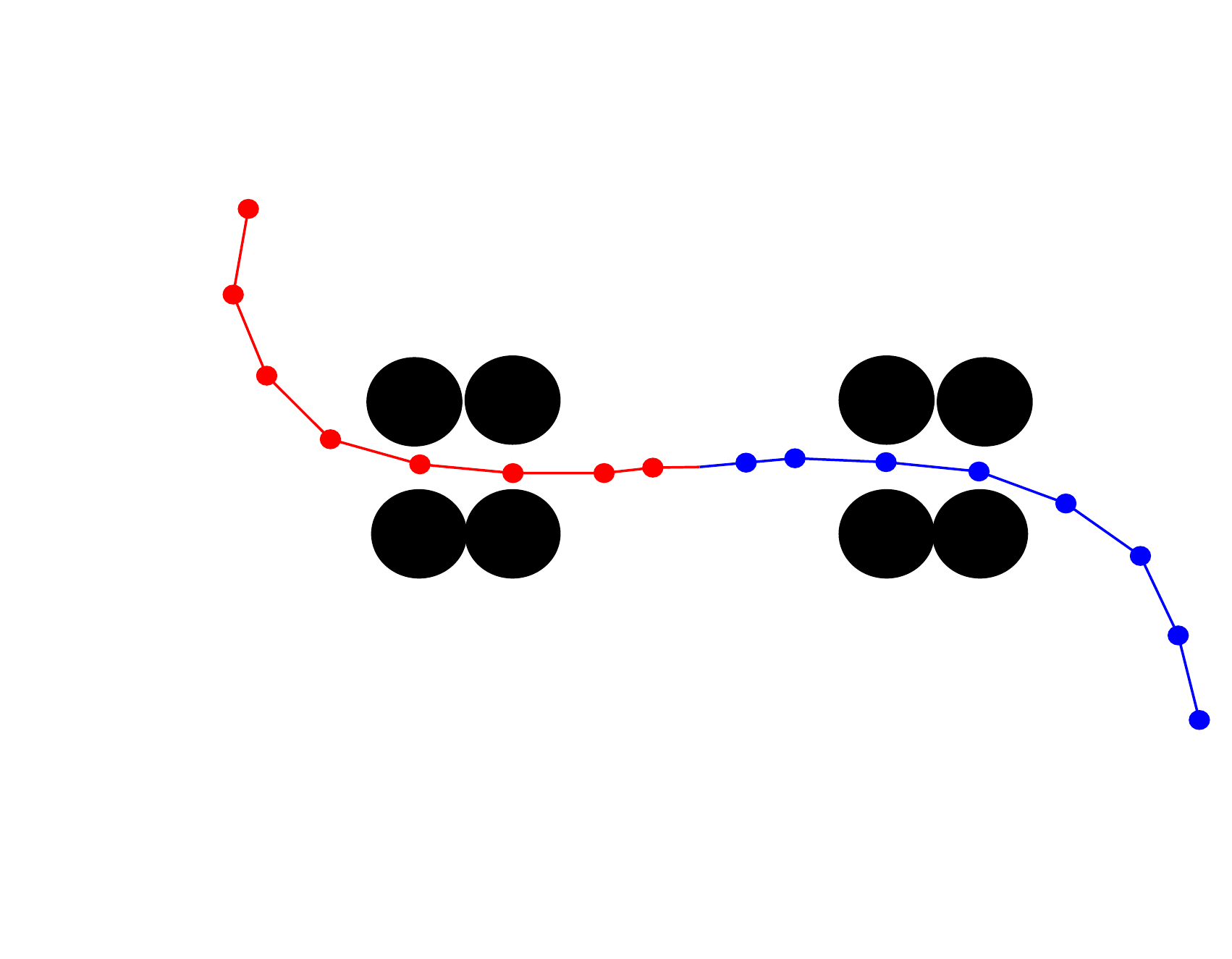}
    \end{subfigure}
    \begin{subfigure}[b]{0.49\columnwidth}
        \includegraphics[width=\linewidth]{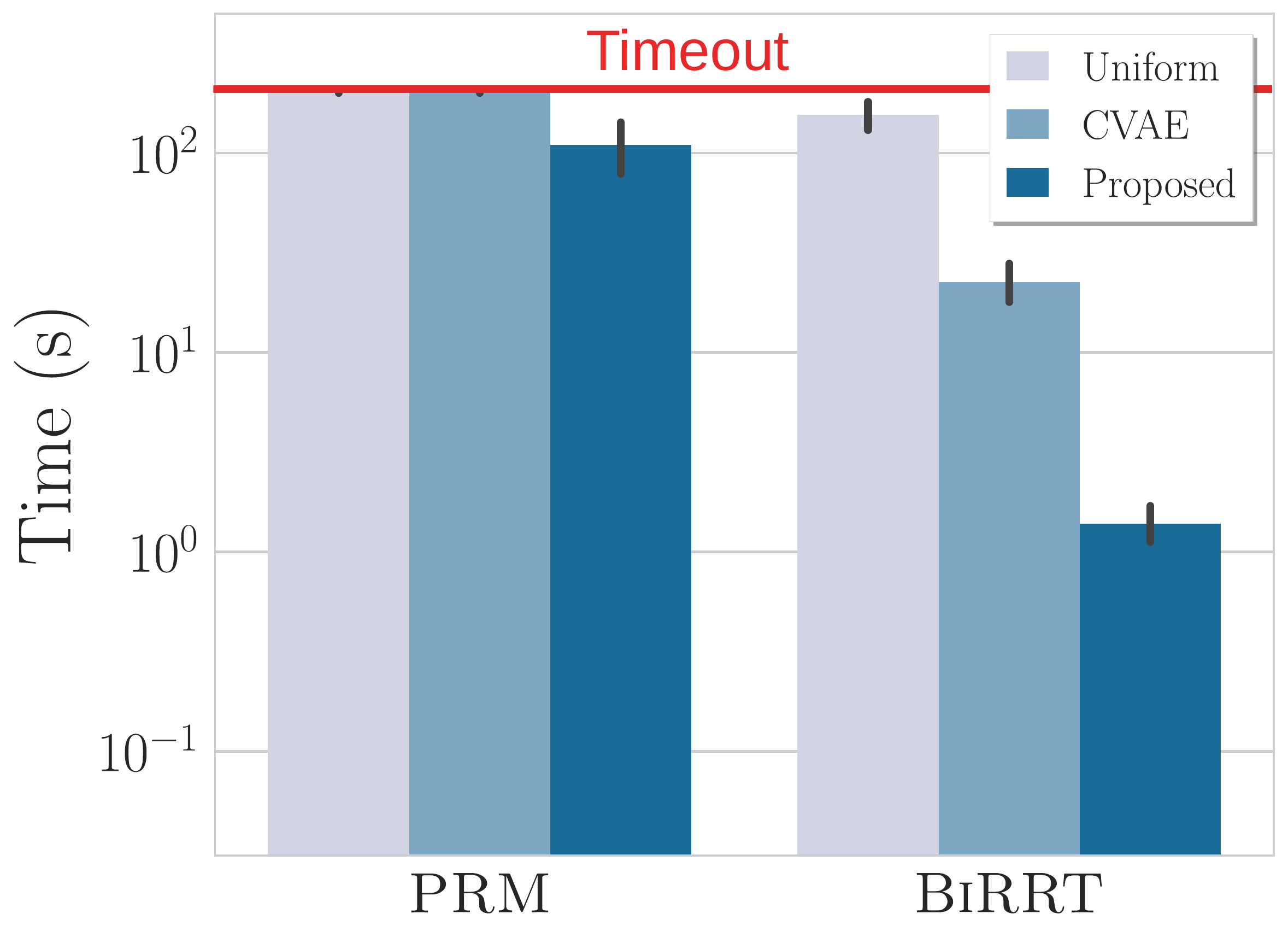}        
    \end{subfigure}
    \caption{Scenario with large decomposition error.}
    \label{fig:sce2}
\end{figure}
The second scenario \autoref{fig:sce2}, has a large \textit{decomposition error} due to the proximity of the local primitives. Each local sampler is trained only on a pair of circles, and thus the majority of the samples produced are invalid when other local primitives are nearby. However, it can be seen that our method still significantly outperformed the other ones. Most notably \textsc{birrt}, with our method, found solutions in 1s while the \textsc{cvae} needed around 20s and the Uniform around 200s. Note also that \mbox{Uniform-\textsc{prm}} and \mbox{\textsc{cvae}-\textsc{prm}} \mbox{timed-out} in all cases. This scenario shows that the proposed method is very robust to approximation errors and can potentially work on very complicated environments that are decomposed into circles.

\subsubsection{Scenario3}
\begin{figure}    
    \begin{subfigure}[b]{0.49\columnwidth}
        \includegraphics[width=\linewidth]{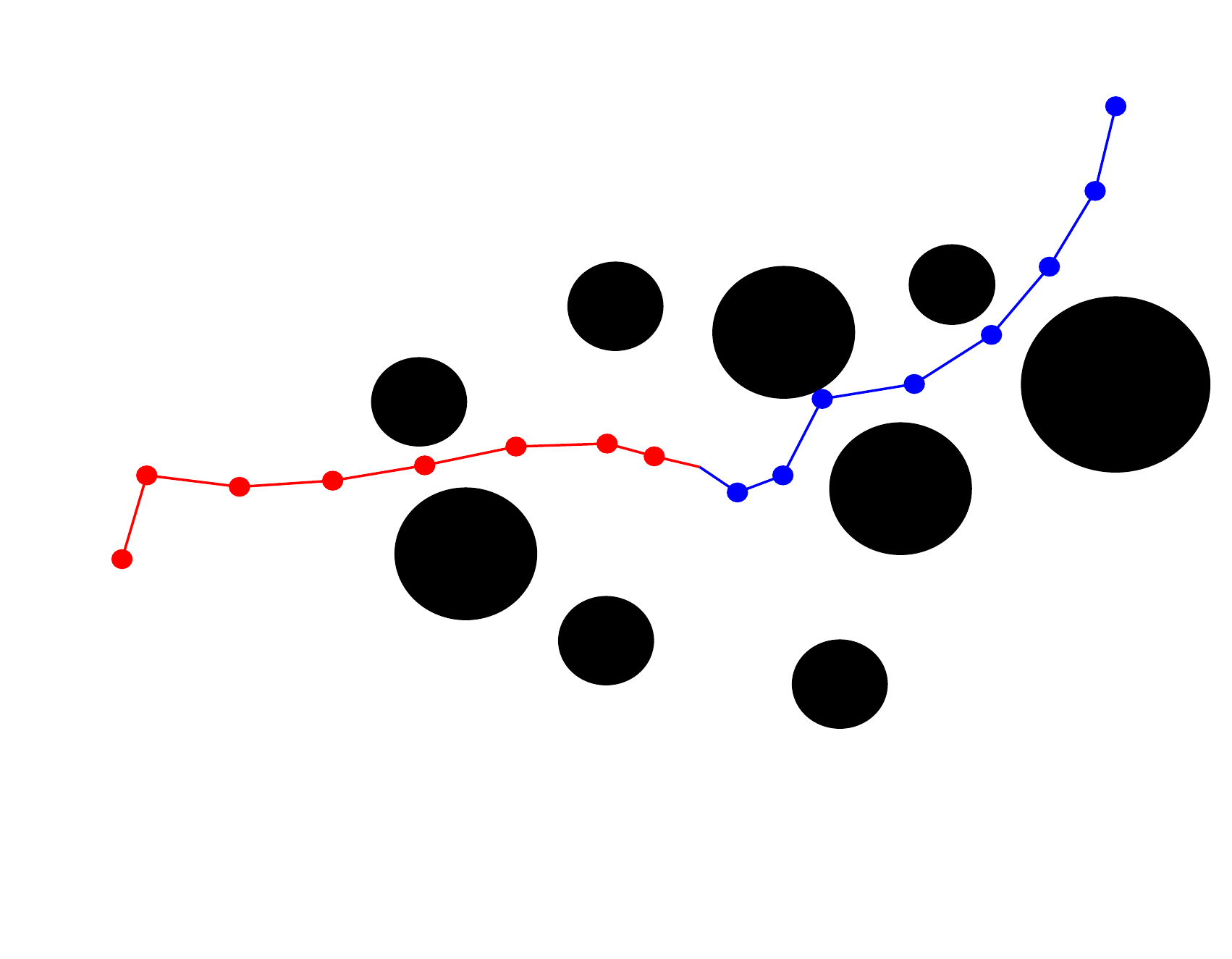}
    \end{subfigure}
    \begin{subfigure}[b]{0.49\columnwidth}
        \includegraphics[width=\linewidth]{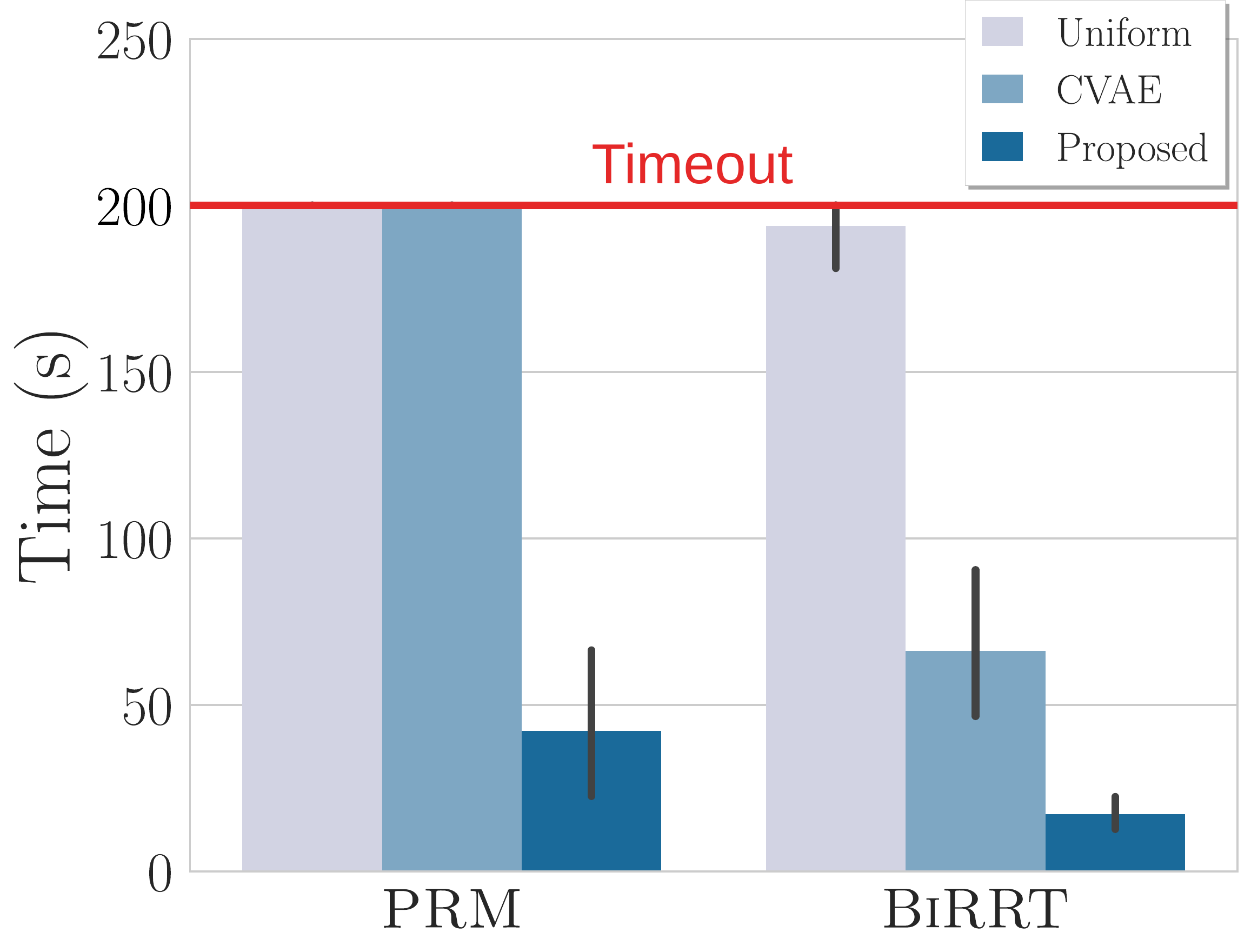}
    \end{subfigure}
    \caption{Scenario with large decomposition and similarity error.}
    \label{fig:sce3}
\end{figure}

This scenario~\autoref{fig:sce3}, is similar to the one used in~\cite{Zucker2008} but much more difficult due to the closeness of the obstacles and the relatively large size of the robot. Both the \textit{decomposition error} as well as  the \textit{similarity error} are large. In~\autoref{fig:pipeline} the different local primitives and their local samplers which were used to create the global sampler are visible. Note that one circle is not part of any local primitive because it is not close to any other circle.
Also in this scenario, our method outperformed the others with \textsc{prm} succeeding only when using our method and \textsc{birrt} needing more than 10 minutes to find a solution with uniform sampling.

\subsection{8-DOF Robot}
\begin{figure}
    \centering
    \begin{subfigure}[b]{0.9\columnwidth}
        \includegraphics[width =1\linewidth]{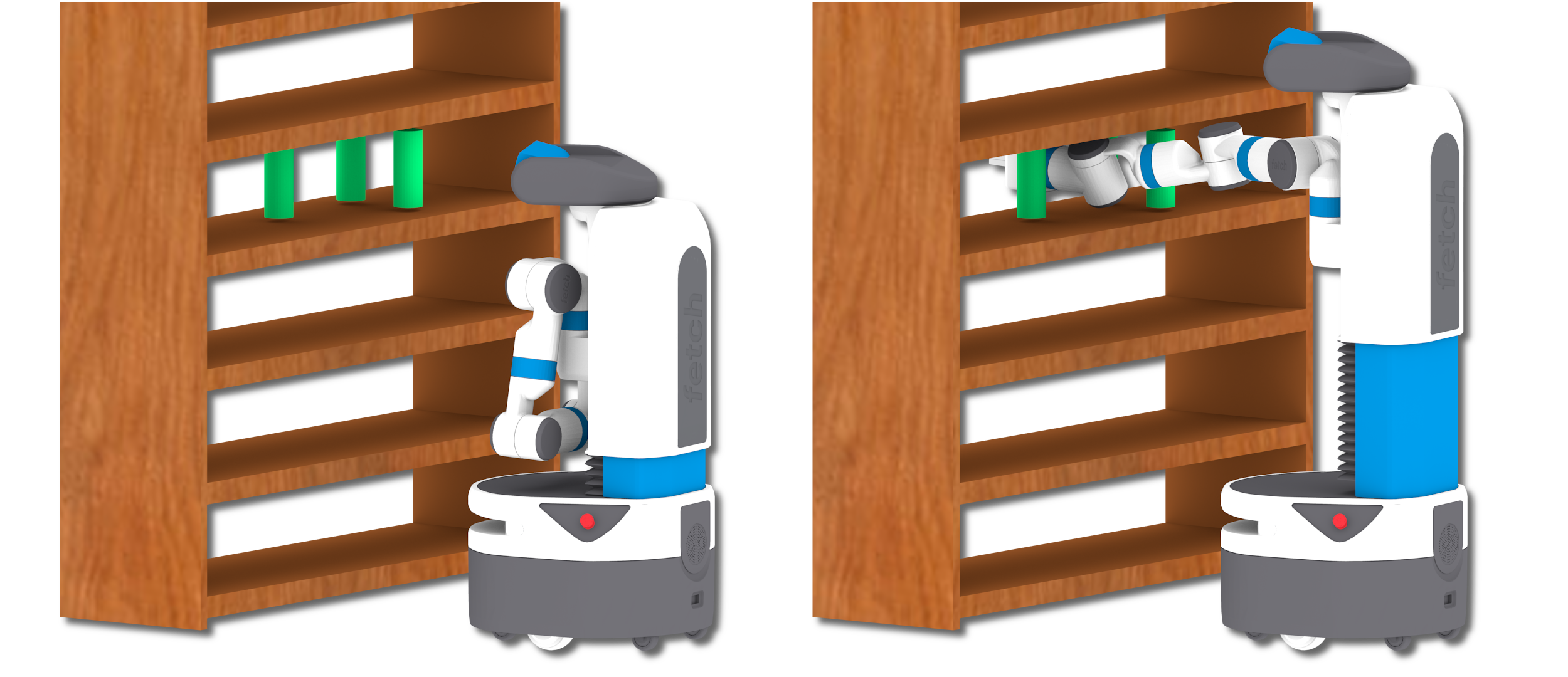}

        \caption{Start Configuration (Left), Goal Configuration (Right).}
        \label{fig:fetch2}

        \vspace{7pt}
    \end{subfigure}
     \begin{subfigure}[b]{0.49\columnwidth}
        \includegraphics[width=\linewidth]{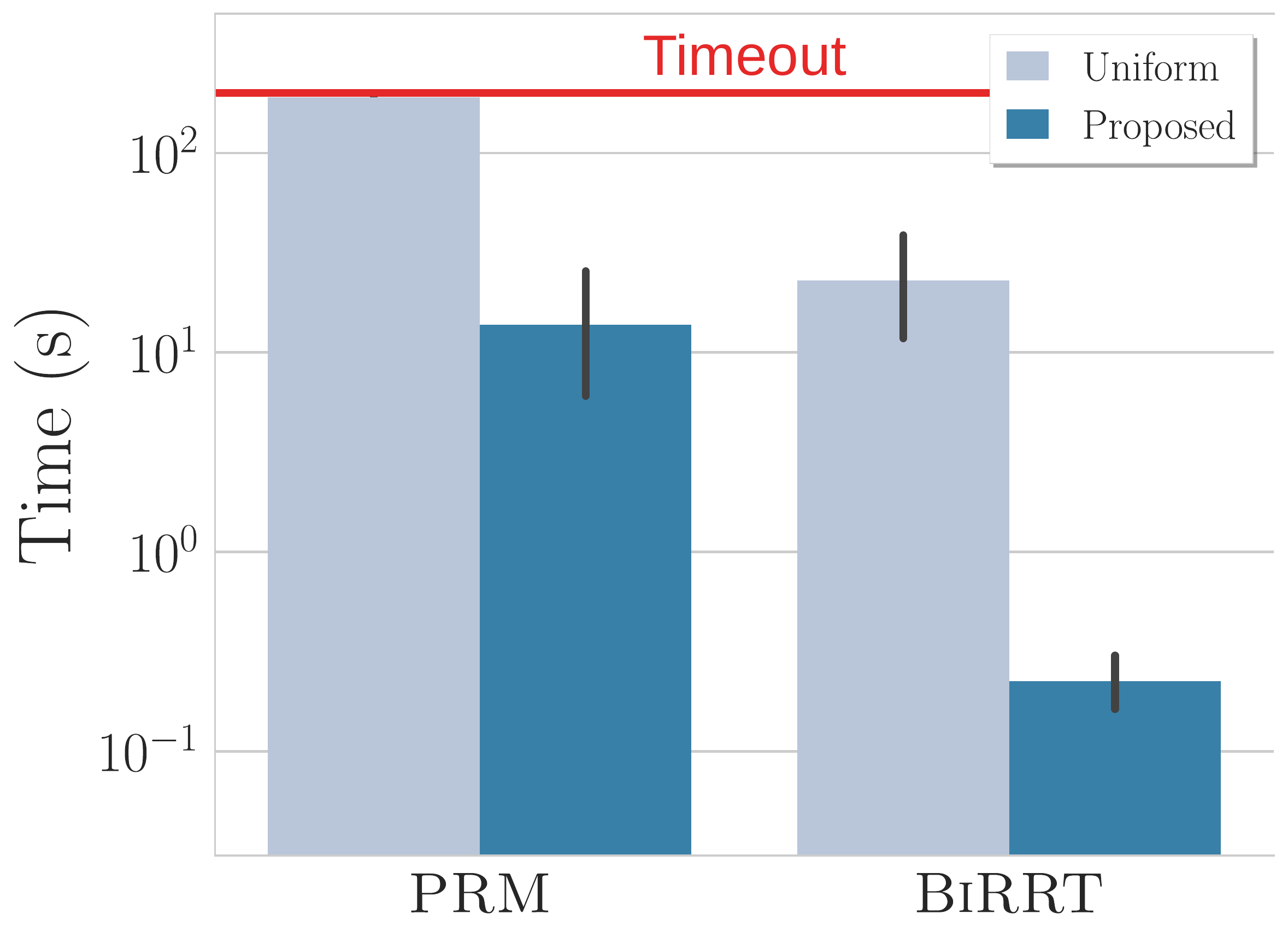}

        \vspace{-5pt}
        \caption{Thin cylinders scenario.}
        \label{fig:fetchsce1}
    \end{subfigure}
    \begin{subfigure}[b]{0.49\columnwidth}
        \includegraphics[width=\linewidth]{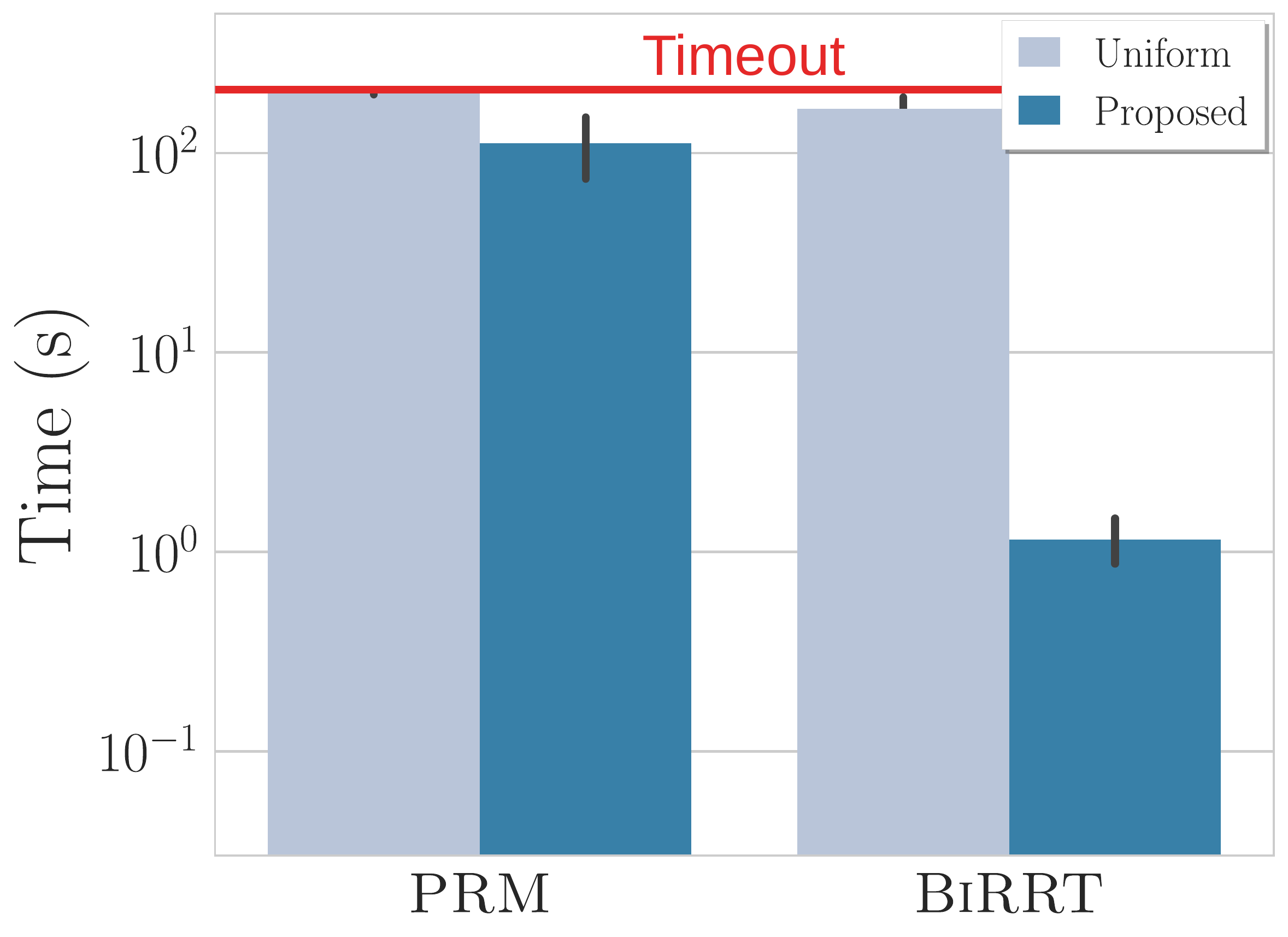}

        \vspace{-5pt}
        \caption{Thick cylinders scenario.}
        \label{fig:fetchsce2}
    \end{subfigure}
    \caption{Challenging motion planning problem and results for the Fetch Robot.} 
\end{figure}

We also experimented on a simulated Fetch Robot~\cite{Wise2016} performing an object manipulation task. The robot has a \mbox{7-DOF} arm and a moving torso resulting in an \mbox{8-DOF} \mbox{C-Space}.
The Fetch Robot tries to place its arm between the cylinders with start and goal shown in \autoref{fig:fetch2}, which is a difficult problem as it requires reaching into a deep shelf.

To demonstrate the practicality of this approach, we constructed a small database using only the local primitives that were present in the test scene. These local primitives were 3 pairs, each with one of the cylinders and the bookcase. Note that since each local primitive contains only 1 cylinder, it is significantly easier to solve than the full problem. We tested the proposed method on 2 scenarios. The first scenario has the same local primitives that existed in the database thus a \textit{similarity error} of zero but a high \textit{decomposition error}. The second scenario has thicker  cylinders with a double radius which introduces a \textit{similarity error}.

We only benchmarked against uniform sampling since there was not a rich dataset to train the \textsc{cvae}. In the results, the increased difficulty of the second  scenario is clearly visible. Both planners had zero or low success rate with uniform sampling while our approach succeeded in all of them. 
\section{CONCLUSIONS}
In this work, we proposed a new sampling-biasing framework that is based on a decomposition of the workspace. We only considered simple geometric primitives, yet we solved problems that were either not possible to solve with the methods we considered or not efficiently solved.
Although we consider our results preliminary, we believe that this work paves a new way to apply experience in motion planning problems.
Future work could include the use of more complicated primitives that are general and can effectively decompose any workspace. Finally, as the database grows in size efficient real-time retrieving algorithms should be used.

\section*{ACKNOWLEDGMENTS}
The authors thank B.~Willey, J.~Hernandez, J.~Abella, and M.~Moll for their valuable and interesting conversations. The authors especially thank Z.~Kingston for the visualization and benchmarking tools that made the Fetch experiment possible.
\addtolength{\textheight}{-9cm}   



\bibliographystyle{ieeetr}
\bibliography{biblio.bib}

\begin{thebibliography}{10}

\bibitem{Dantam2018}
N.~T. Dantam, Z.~K. Kingston, S.~Chaudhuri, and L.~E. Kavraki, ``An incremental
  constraint-based framework for task and motion planning,'' {\em The
  International Journal of Robotics Research (IJRR)}, vol.~37, no.~10,
  pp.~1134--1151, 2018.

\bibitem{Choset2005}
H.~M. Choset, S.~Hutchinson, K.~M. Lynch, G.~Kantor, W.~Burgard, L.~E. Kavraki,
  and S.~Thrun, {\em Principles of Robot Motion: Theory, Algorithms, and
  Implementation}.
\newblock MIT press, 2005.

\bibitem{Hsu2003a}
D.~Hsu, T.~Jiang, J.~Reif, and Z.~Sun, ``{The bridge test for sampling narrow
  passages with probabilistic roadmap planners},'' {\em IEEE International
  Conference on Robotics and Automation (ICRA)}, vol.~3, pp.~4420--4426, 2003.

\bibitem{Hsu1998}
D.~Hsu, L.~E. Kavraki, J.-C. Latombe, R.~Motwani, S.~Sorkin, {\em et~al.}, ``On
  finding narrow passages with probabilistic roadmap planners,'' in {\em
  Robotics: The Algorithmic Perspective: Workshop on the Algorithmic
  Foundations of Robotics}, pp.~141--154, 1998.

\bibitem{Amato1998}
N.~M. Amato, O.~B. Bayazit, L.~K. Dale, C.~Jones, and D.~Vallejo, ``{OBPRM: An
  Obstacle-Based PRM for 3D Workspaces},'' in {\em Third Workshop on the
  Algorithmic Foundations of Robotics on Robotics: The Algorithmic
  Perspective}, pp.~155--168, 1998.

\bibitem{Coleman2015}
D.~Coleman, I.~A. Sucan, M.~Moll, K.~Okada, and N.~Correll, ``{Experience-based
  planning with sparse roadmap spanners},'' in {\em IEEE International
  Conference on Robotics and Automation (ICRA)}, pp.~900--905, June 2015.

\bibitem{Ichter2018}
B.~Ichter, J.~Harrison, and M.~Pavone, ``{Learning Sampling Distributions for
  Robot Motion Planning},'' in {\em IEEE International Conference on Robotics
  and Automation (ICRA)}, pp.~7087--7094, May 2018.

\bibitem{Phillips2012}
M.~Phillips, B.~J. Cohen, S.~Chitta, and M.~Likhachev, ``E-graphs:
  Bootstrapping planning with experience graphs,'' in {\em Robotics Science and
  Systems (RSS)}, July 2012.

\bibitem{Kim2017c}
B.~Kim, L.~P. Kaelbling, and T.~Lozano-Perez, ``{Learning to guide task and
  motion planning using score-space representation},'' in {\em IEEE
  International Conference on Robotics and Automation (ICRA)}, pp.~2810--2817,
  2017.

\bibitem{Lehner2018}
P.~Lehner and A.~Albu-Schaffer, ``{The Repetition Roadmap for Repetitive
  Constrained Motion Planning},'' {\em IEEE Robotics and Automation Letters
  (RAL)}, vol.~3, no.~3, pp.~3884 -- 3891, 2018.

\bibitem{Kavraki1996}
L.~Kavraki, P.~Svestka, J.-C. Latombe, and M.~Overmars, ``{Probabilistic
  roadmaps for path planning in high-dimensional configuration spaces},'' {\em
  IEEE Transactions on Robotics and Automation (TRA)}, vol.~12, no.~4, pp.~566
  -- 580, 1996.

\bibitem{Kuffner2000}
J.~Kuffner and S.~LaValle, ``{RRT-connect: An efficient approach to
  single-query path planning},'' {\em IEEE International Conference on Robotics
  and Automation (ICRA): Millennium Conference}, vol.~2, pp.~995--1001, 2000.

\bibitem{Hsu}
D.~Hsu, J.-C. Latombe, and R.~Motwani, ``{Path planning in expansive
  configuration spaces},'' in {\em International Conference on Robotics and
  Automation (ICRA)}, vol.~3, pp.~2719--2726, 1997.

\bibitem{Canny1988}
J.~Canny, {\em The Complexity of Robot Motion Planning}.
\newblock MIT press, 1988.

\bibitem{Canny1988b}
J.~Canny, ``{Some algebraic and geometric computations in PSPACE},'' in {\em
  Annual ACM symposium on Theory of computing (STOC)}, pp.~460--467, 1988.

\bibitem{Boor1999a}
V.~Boor, M.~H. Overmars, and a.~V.~D. Stappen, ``{The Gaussian sampling
  strategy for probabilistic roadmap planners},'' in {\em IEEE International
  Conference on Robotics and Automation (ICRA)}, vol.~2, pp.~1018--1023, May
  1999.

\bibitem{Brock2004}
O.~Brock, ``{Adapting the sampling distribution in PRM planners based on an
  approximated medial axis},'' in {\em IEEE International Conference on
  Robotics and Automation (ICRA)}, vol.~5, pp.~4405--4410, April 2004.

\bibitem{Zucker2008}
M.~Zucker, J.~Kuffner, and J.~A. Bagnell, ``{Adaptive workspace biasing for
  sampling-based planners},'' {\em IEEE International Conference on Robotics
  and Automation (ICRA)}, pp.~3757--3762, 2008.

\bibitem{Burns2005}
B.~Burns and O.~Brock, ``{Toward optimal configuration space sampling},'' in
  {\em Robotics Science and Systems (RSS)}, pp.~105--112, 2005.

\bibitem{Arslan2015}
O.~Arslan and P.~Tsiotras, ``{Machine learning guided exploration for
  sampling-based motion planning algorithms},'' {\em International Conference
  on Intelligent Robots and Systems (IROS)}, pp.~2646--2652, December 2015.

\bibitem{Denny2013}
J.~Denny and N.~M. Amato, ``{Toggle PRM: A coordinated mapping of C-free and
  C-obstacle in arbitrary dimension},'' {\em Springer Tracts in Advanced
  Robotics}, vol.~86, pp.~297--312, 2013.

\bibitem{Berenson2012a}
D.~Berenson, P.~Abbeel, and K.~Goldberg, ``{A robot path planning framework
  that learns from experience},'' {\em International Conference on Robotics and
  Automation (ICRA)}, pp.~3671--3678, 2012.

\bibitem{Lien2009}
J.-M. Lien and Y.~Lu, ``{Planning motion in environments with similar
  obstacles.},'' {\em Robotics Science and Systems (RSS)}, 2009.

\bibitem{Jetchev2013}
N.~Jetchev and M.~Toussaint, ``Fast motion planning from experience: trajectory
  prediction for speeding up movement generation,'' {\em Autonomous Robots},
  vol.~34, no.~2, pp.~111--127, 2013.

\bibitem{Geraerts2007}
R.~Geraerts and M.~H. Overmars, ``{Creating high-quality paths for motion
  planning},'' {\em International Journal of Robotics Research (IJRR)},
  vol.~26, no.~8, pp.~845--863, 2007.

\bibitem{sucan2012}
I.~A. {\c{S}}ucan, M.~Moll, and L.~E. Kavraki, ``The {O}pen {M}otion {P}lanning
  {L}ibrary,'' {\em {IEEE} Robotics \& Automation Magazine}, vol.~19,
  pp.~72--82, December 2012.

\bibitem{moll2015}
M.~Moll, I.~A. {\c{S}}ucan, and L.~E. Kavraki, ``Benchmarking motion planning
  algorithms: An extensible infrastructure for analysis and visualization,''
  {\em {IEEE} Robotics \& Automation Magazine}, vol.~22, pp.~96--102, September
  2015.

\bibitem{Wise2016}
M.~Wise, M.~Ferguson, D.~King, E.~Diehr, and D.~Dymesich, ``Fetch and freight:
  Standard platforms for service robot applications,'' in {\em Workshop on
  Autonomous Mobile Service Robots}, 2016.

\end{thebibliography}

\end{document}